# MULTIBAND SAS IMAGERY

Isaac D Gerg    Applied Research Laboratory, Pennsylvania State University, USA

## 1    INTRODUCTION

Advances in unmanned synthetic aperture sonar (SAS) imaging platforms allow for the simultaneous collection of multiband SAS imagery.  The imagery is collected over several octaves and the phenomenology's interactions with the sea floor vary greatly over this range -- higher frequencies resolve proud & fine structure of the seafloor while lower frequencies resolve subsurface features and often induce internal resonance in man-made objects.

Currently, analysts examine multiband imagery by viewing a single band at a time.  This method makes it difficult to ascertain correlations between any pair of bands collected over the same location.   To mitigate this issue, we propose methods which ingest high frequency (HF) and low frequency (LF) SAS imagery and generates a color composite creating what we call a multiband SAS (MSAS) image.  The MSAS image contains the relevant portions of the HF and LF images required by an analyst to interpret the scene and are defined using a spatial saliency metric computed for each image.  We then combine the saliency and acoustic backscatter measures to form the final MSAS image.

We investigate three fusion schemes. The first two schemes -- one based on a constant false alarm rate (CFAR) detector and one based on speeded up robust features[1] (SURF) densities -- fuse the data in a human visual system (HVS) focused color space CIELAB[2] while the third scheme fuses by using dual colormaps -- one for salient HF features and one for salient LF features.  We evaluate our results by examining three similarity metrics on the original images and the fused image. The metrics we examined are structural similarity index metric[3] (SSIM), normalized cross correlation (NCC), and mean-squared-error (MSE).

We demonstrate our techniques using imagery collected from a dual band SAS platform consisting an HF and LF band existing two-and-a-half octaves apart.  The imagery was collected over seafloors of medium sand containing rocks and ripples. The images input to our algorithms are normalized to [0 1] domain and are post-processed to be human consumable by removing range-varying gain and are dynamic range compressed using an algorithm similar to the rational mapping operator of Schlick[4].

We especially examine one set of images, which we call the "ripple" dataset, to demonstrate our fusion algorithms on a natural scene.  Figure 1 shows the ripple dataset along with its joint and marginal probability density functions (pdfs).

The author gratefully acknowledges the US Office of Naval Research for its support of this work (Contract  Number: N00014-10-G-0259)
**Vol. 36. Pt.1 2014**

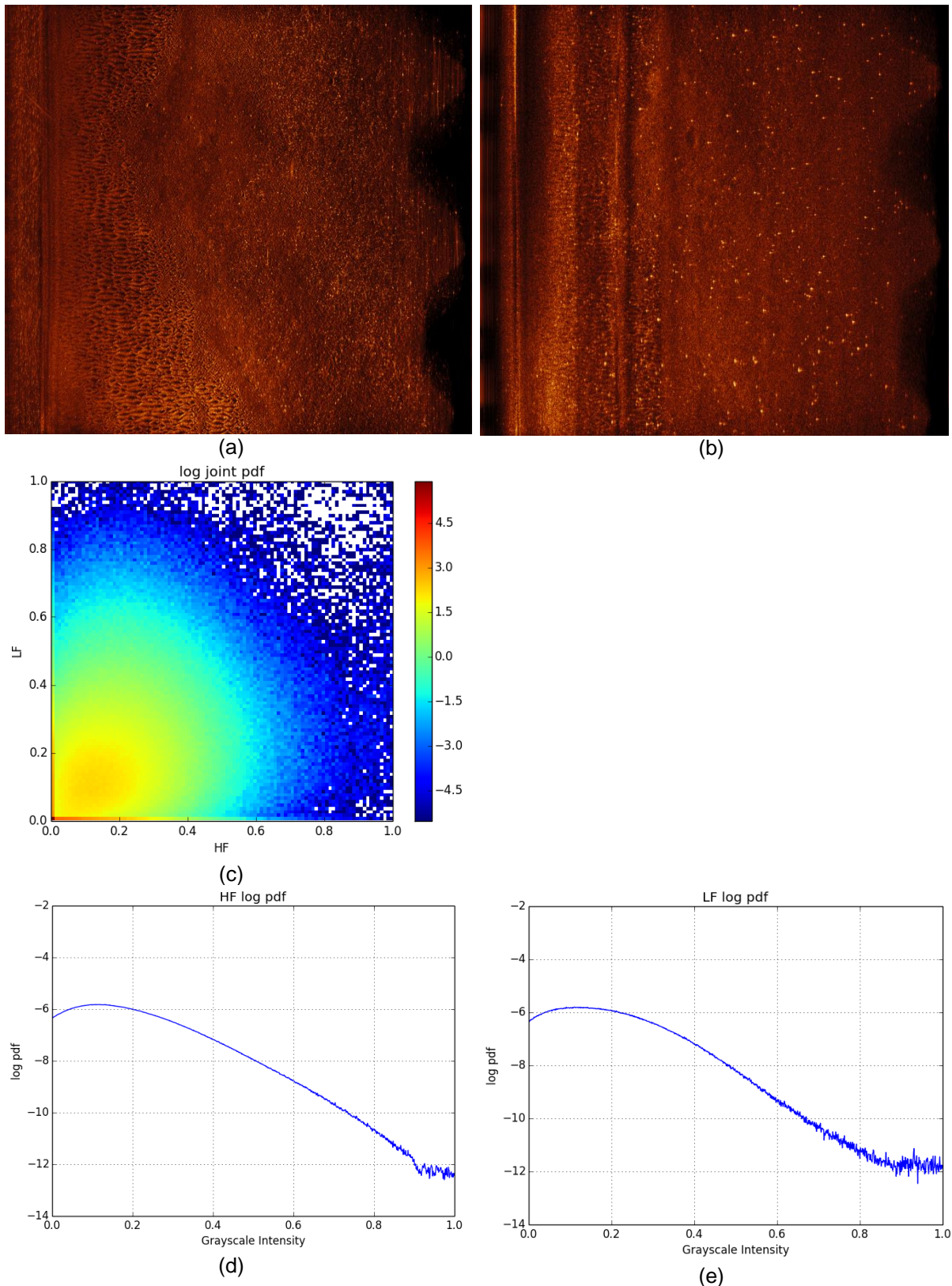

**Figure 1** (a), (b) HF/LF (respectively) images of the ripple dataset. (c) log joint pdf of HF/LF image pair. (d), (e) log pdf of HF and LF (respectively) images .

## 2 CIELAB FUSION SCHEMES

### 2.1 Introduction to CIELAB Color Space

Canonical color spaces such as red-green-blue (RGB) or hue-saturation-value (HSV) decompose color into a basis suitable for computer display – they are largely driven by the RGB, specifically sRGB[1], color space. Such color spaces fail to account for the effects of human perception. For example, a red object and a blue object with identical surface reflectance under ordinary illumination are perceived as different lightness by the HVS with the red object appearing darker than the blue object. Neither RGB nor HSV color spaces account for this property.

CIELAB color space was designed to account for the nonlinearities of the HVS by relating color differences to perceived human sensation. Moving a fixed distance in any direction from a starting point yields the same perceived difference in color independent of the direction moved. Such a color space is referred to as *perceptually uniform*. Using the previous example of the red and blue objects under the same illumination and observation geometry, CIELAB is able to characterize the perceived differences in illumination where RGB or HSV cannot.

### 2.2 Gamut Considerations When Transforming from CIELAB to sRGB

CIELAB color space contains all perceivable colors and is independent of display device. However, a subset of colors representable in CIELAB cannot be represented in sRGB. The valid subspace, referred to as gamut, must be taken into account when performing operations in CIELAB and transforming the results to sRGB. The easiest method to account for out-of-gamut colors is simply to force colors outside the gamut to the maximum the gamut supports. This method is known as *gamut clipping* and can result in perceivable distortion causing the naturalness of the image to be lost.

We performed an optimization which scales the chroma in CIELAB to a point at which all the colors are in the sRGB subspace to mitigate gamut clipping when converting from CIELAB to sRGB. We found this optimization to significantly increase computation time. We alleviate this computation burden by reducing gamut clipping by restricting CIELAB values to a subspace mostly overlapping with the sRGB subspace. We find this method produces acceptable image quality. Specifically, we reduce our gamut by limiting the *chroma*, the amount of saturation exhibited by a color. In CIELAB, chroma is defined as the distance from the origin to a color point projected onto the a\*-b\* plane and hue is defined as the angle of this vector.

When converting from CIELAB to sRGB, some hues have more chroma headroom than others for a given luminance due to the nonlinearities of the HVS and the design of sRGB. Because of this and the fact the histograms of the original images have a mode near the lower end of the intensity scale, we choose hues in the fourth quadrant of the a\*-b\* plane. Figure 2 shows a slice of valid sRGB space in CIELAB for a given luminance[5].

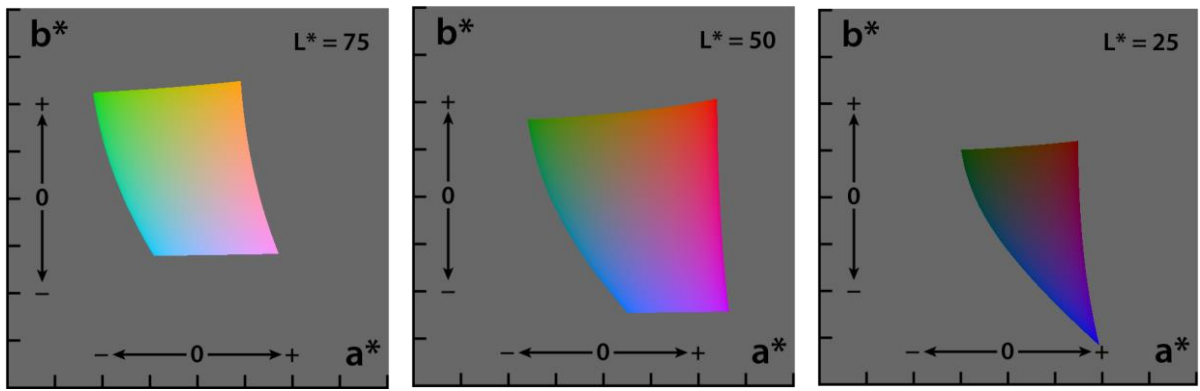

**Figure 2. Valid sRGB gamut in CIELAB shown for luminance L\*=75,50,25 (a), (b), (c) respectively. Notice as L\* gets smaller, the valid sRGB gamut moves to the fourth quadrant of a\*-b\*.**

### 2.3   Structure of CIELAB Fusion Algorithms

For these fusion algorithms, we divide the final image content into two components: luminance and color. For each pixel in the output image, the luminance is derived from the acoustic intensity and the color from its saliency.

### 2.4   Development of CIELAB Fusion Algorithm Using CFAR Saliency Metric

For this technique, we assume everything in the HF image is salient since it is of high resolution and represents proud features – generally useful to an analyst. Saliency for LF is computed by subtracting the image from a low pass filtered version of itself. The resulting image is then thresholded whereby values less than the threshold are set to zero and values above the threshold are untouched.

$$f' = f - f_{background} \quad (1)$$
$$f_{salient} = \begin{cases} f' & f' > \alpha \\ 0 & otherwise \end{cases} \quad (2)$$

where $f_{background}$ is a low-pass filtered version of *f* which provides an estimate of the background level. The threshold $\alpha$ is used to determine if a particular pixel is salient relative to the background. The similarities of this algorithm to a CFAR detector give rise to calling it the CFAR saliency metric.

We compute the luminance channel by combining the HF image and the salient LF image using the supremum function.

$$L^* = \sup(f_{salient}, f_{HF}) \quad (3)$$

The fused image pixel color is determined by LF saliency. The colors sweep from $hue_0$ to $hue_1$ where $hue_0$ is used when an LF pixel has no saliency and $hue_1$ is used when an LF pixel is completely salient. We fix the chroma for the pixel to $chroma_0$ so that pixels in the a\*-b\* plane lie along a smooth manifold that is mostly in the sRGB gamut.

The idea of using a fixed chroma is an important one. If the chroma is a free parameter, the resulting MSAS image contains ambiguities because it results in pixels with low saturation (i.e. gray colors) destroying the ability to determine what band the pixel is salient .

In our implementation, we set $\alpha = 0.0$ and bound the hue to lie in the fourth quadrant of a\*-b\*. We estimate $f_{background}$ using a boxcar filter over a five meter area.

### 2.5 Development of CIELAB Fusion Algorithm using SURF Features

For this technique, we compute saliency maps of both HF and LF images. The fused image pixel luminance is computed as a weighted average of the HF and LF image where the weight is determined by the output of the nonlinear mapper based on the relative saliency between bands. The nonlinear mapping function derives the luminance from the HF image unless the LF pixel is salient and the HF pixel is not. Figure 3e depicts this function.

The saliency maps are computed from the density of SURF features derived from a despeckled versions of each image. The densities are then smoothed with a Gaussian kernel. The maps are then fed to the nonlinear mapper function determining the weight, $w$, of the HF and LF image to assign to the luminance channel given by the equation:

$$L^* = f_{HF} w + f_{LF}(1-w) \tag{4}$$

The chroma is fixed and the hue is determined as a function of the LF saliency map, $s_{LF}$, where low saliency is mapped to $hue_0$ and high saliency is mapped to $hue_1$. This is similar to the mapping used in the previous section.

$$\theta = \frac{\pi}{2}(1 - s_{LF}) \tag{5}$$
$$a^* = \cos(\theta)\, C_a \tag{6}$$
$$b^* = -\sin(\theta)\, C_b \tag{7}$$

where $C_a, C_b$ are the maximum chroma allowed for a*, b* respectively.

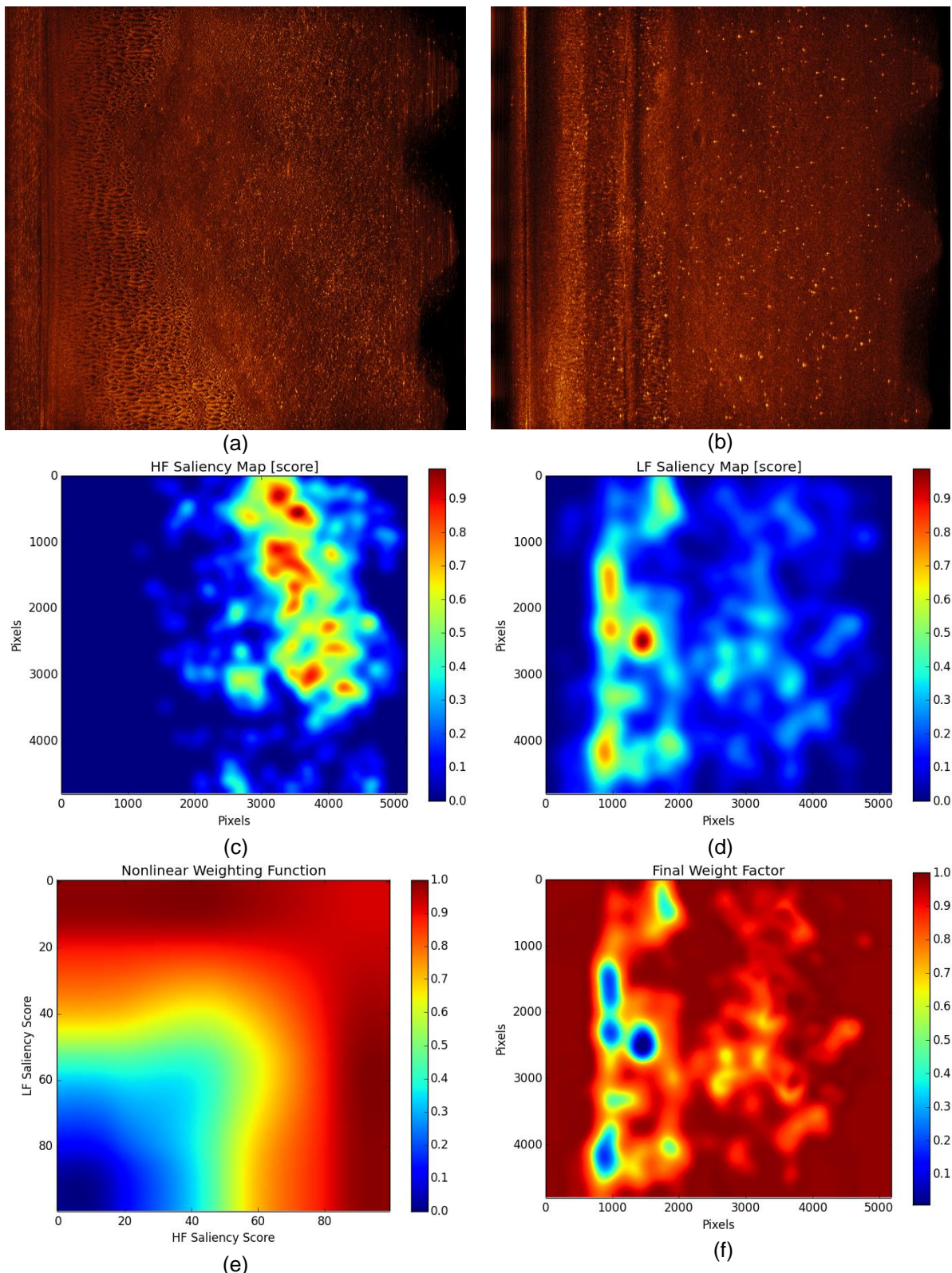

(a) (b) (c) (d) (e) (f)

**Figure 3.** (a), (b) HF, LF images of ripple dataset. (c) (d) HF, LF saliency maps from original images. (c) Nonlinear mapping function used to determine weighted average of luminance channel. (d) Final weight factor, $w$, used in create L* channel.

## 3  DUAL COLORMAP FUSION SCHEME

### 3.1  Development of Dual Colormap Fusion Algorithm Using CFAR Saliency Metric

Our dual color map fusion techniques represents the relevant acoustic intensity features in the luminance channel and represents their derivative sonar band as color just as in the CIELAB fusion algorithms previously mentioned.  However, wherein the CIELAB fusion algorithms the color was determined through algebra in the LAB color space, color here is computed from two predetermined colormaps.

The two colormaps chosen represent HF and LF features respectively.  The colormaps must be perceptually linearly luminant (or nearly so) as to preserve the relative acoustic intensity perceived by the viewer.

The luminance of a pixel is determined by the supremum of the HF and LF intensity images. The colormap to derive each output pixel is determined by the CFAR saliency metric.  When the saliency metric is greater than the threshold, we use the LF colormap to colorize the pixel otherwise the HF colormap is used.

We found several colormap pairings to yield results pleasing to the eye and also naturally interpretable but we preferred one scheme overall: a linearly luminant grayscale colormap for HF and a "hot" colormap for LF.  An example result from this scheme is shown in **Error! Reference source not found.**e.

In our implementation, we set $\alpha = 0.5$ estimate $f_{background}$ using a boxcar filter over a five meter area.

## 4  FUSION PERFORMANCE EVALUATION

. A well fused image has sufficient similarity with both its derivative images such that its information content is more than either source exclusively. We evaluate how well the fused image represents each derivative image by measuring the similarity between the two.  Specifically, we convert the fused image to a perceptually linear grayscale (i.e. the luminance channel of the image in LAB color space) and evaluate its similarity to each HF and LF derivative image. We use three metrics to evaluate our results: normalized cross correlation (NCC), mean squared error, and a perceptual based metric known as the structural similarity index metric (SSIM).  Of the three metrics, the SSIM metric is best matched to the visual human system but the results from the other metrics are presented as traditional means of measuring image similarity

The normalized cross correlation (NCC) is computed as $NCC = \frac{1}{N}\sum \frac{(f-\bar{f})(g-\bar{g})}{\sigma_f \sigma_g}$. The mean squared error (MSE) is computed as $MSE = \frac{1}{N}\sum (g-f)^2$. The SSIM metric is defined as $SSIM = \frac{(2\mu_f \mu_g + c_1)(2\sigma_{fg} + c_2)}{(\mu_f^2 + \mu_g^2 + c_1)(\sigma_f^2 + \sigma_g^2 + c_2)}$.

## 5 RESULTS

### 5.1 Example Results from Ripple Dataset

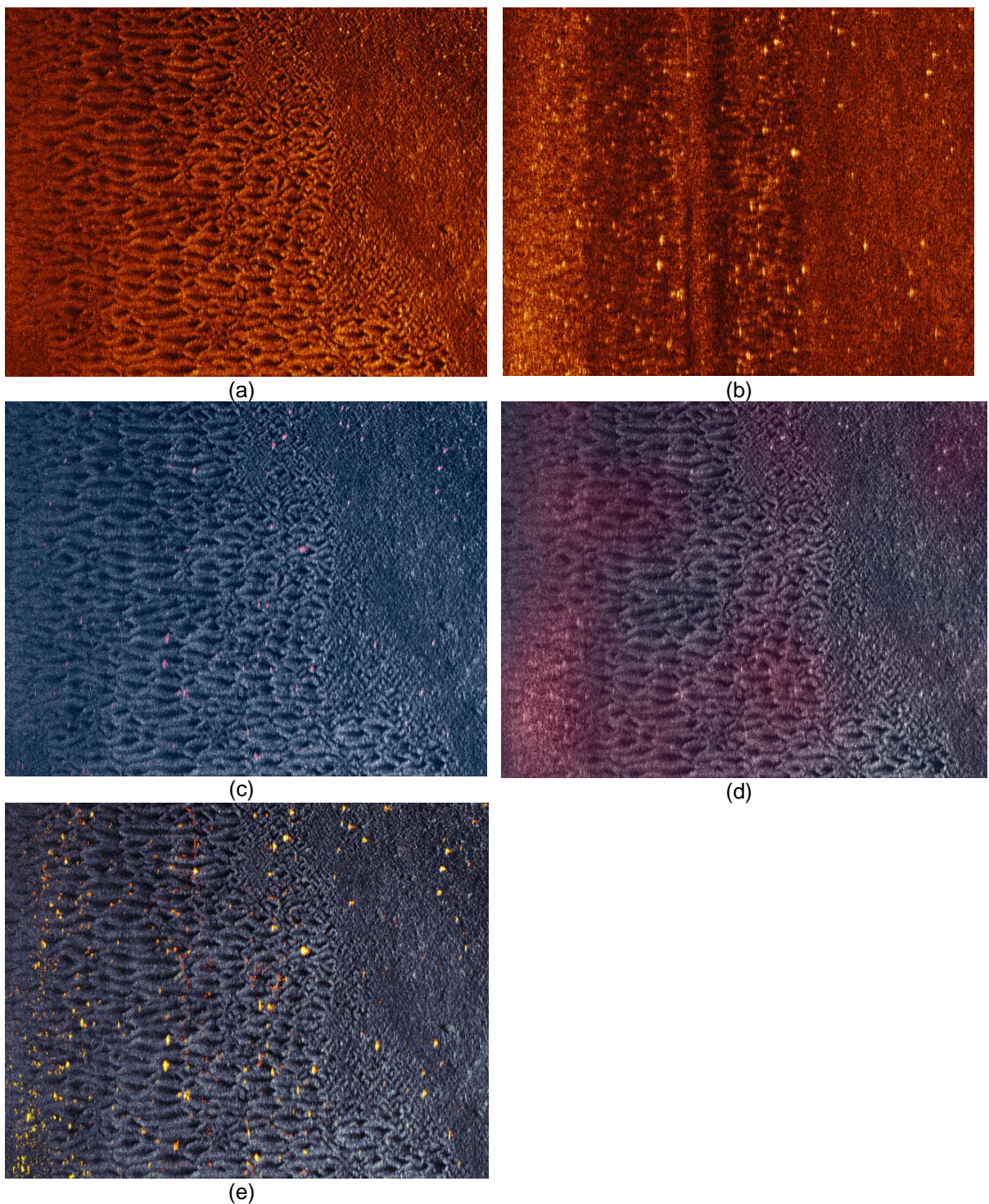

(a)  (b)
(c)  (d)
(e)

**Figure 4. Original images and fused results of the ripple dataset. (a) HF snippet, (b) LF snippet, (c) CFAR saliency fusion, (d) SURF density saliency fusion, (e) Dual colormap fusion**

## 5.2 Fusion Results from an Image Database

We evaluate the fusion algorithms on a database of 264 HF/LF SAS image pairs and measure the CC, MSE, and SSIM. The baseline measurement computes the metric using the HF and LF images. For the SSIM and correlation coefficient metrics (normalized cross correlation), high numbers are better. For the MSE metric, lower numbers are better.

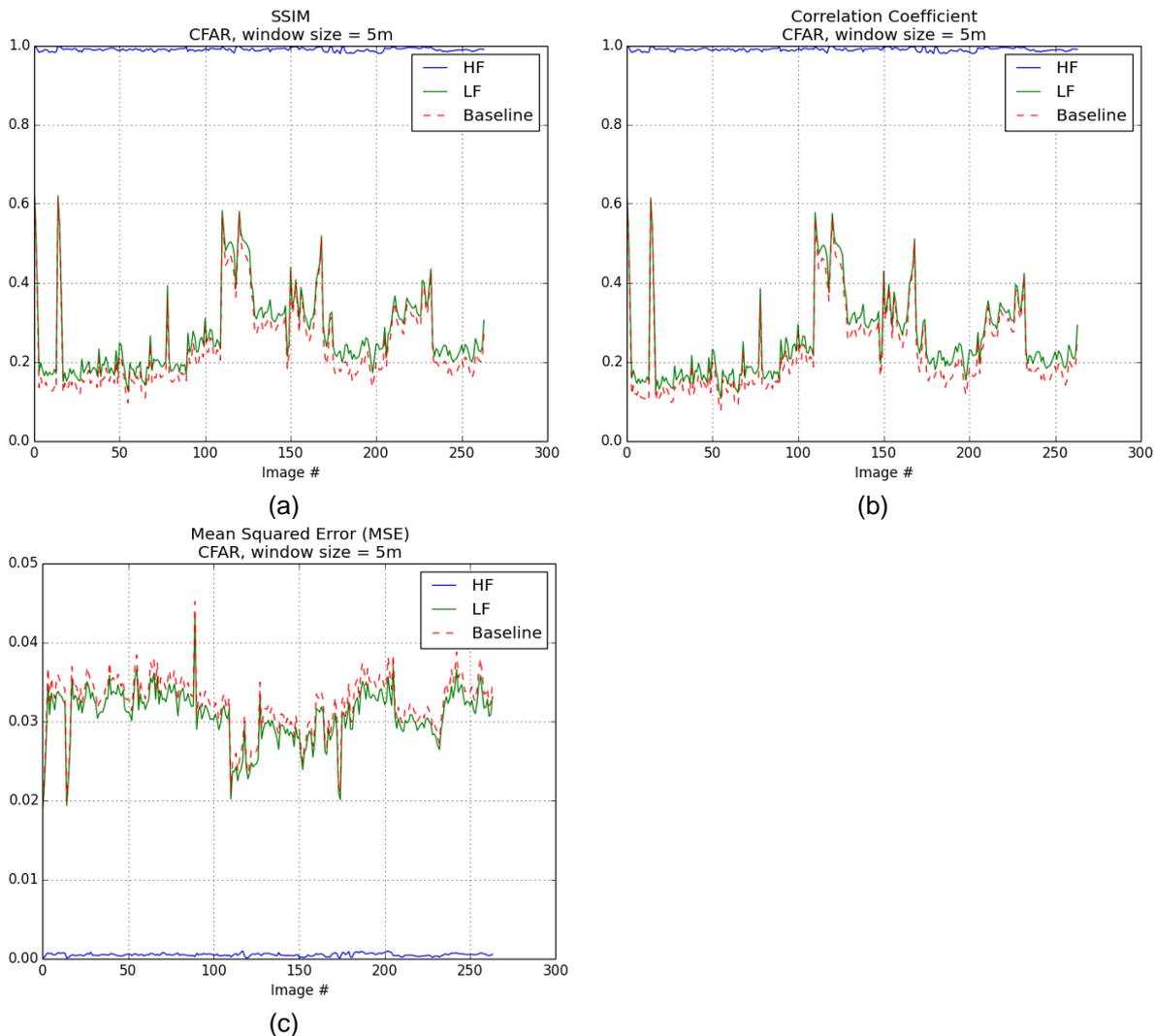

**Figure 5. (a), (b), (c) Images metrics of the CFAR saliency fusion algorithm for SSIM, correlation coefficient, and mean squared error respectively.**

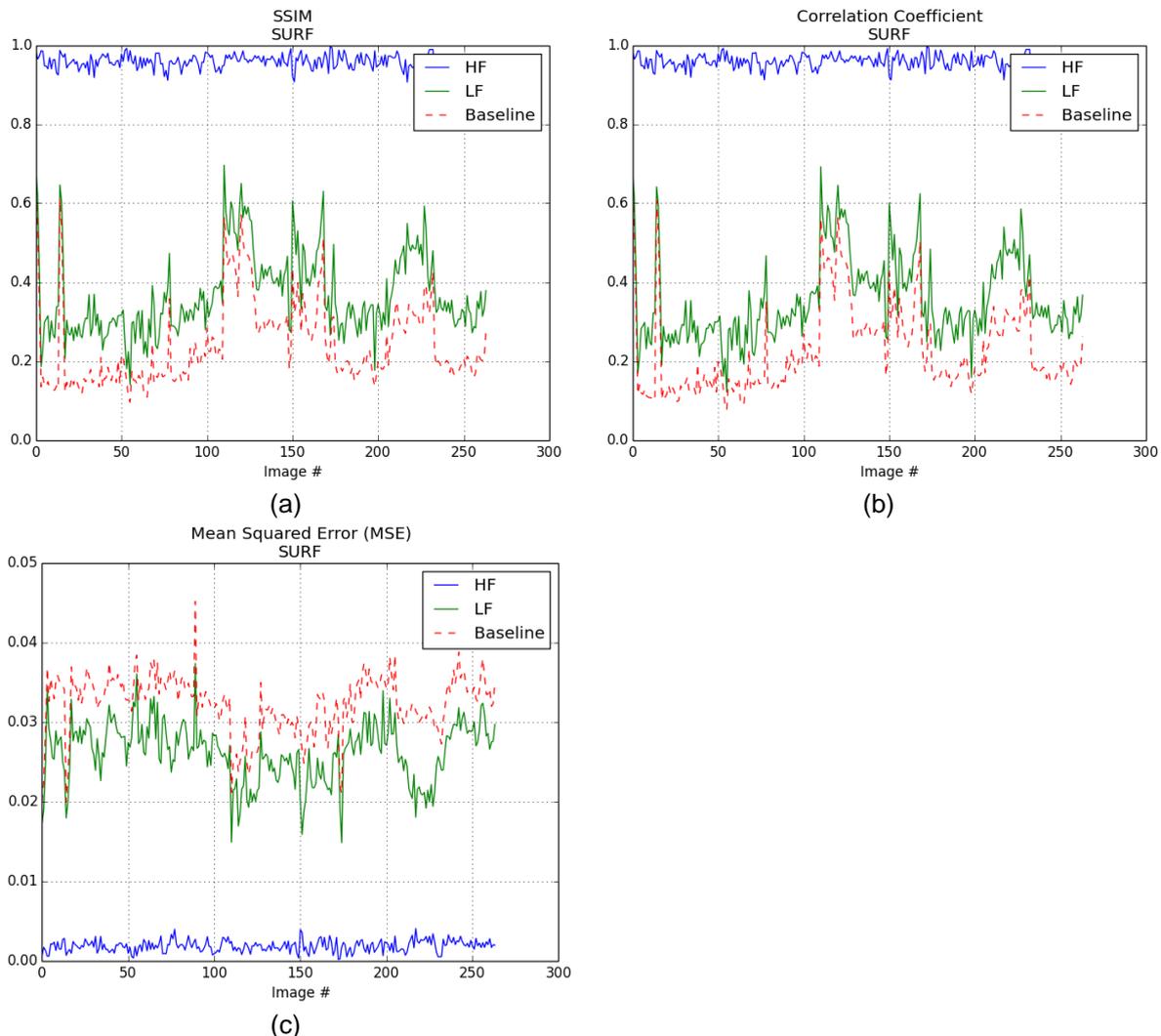

**Figure 6. (a), (b), (c) Images metrics of the SURF saliency fusion algorithm for SSIM, correlation coefficient, and mean squared error respectively.**

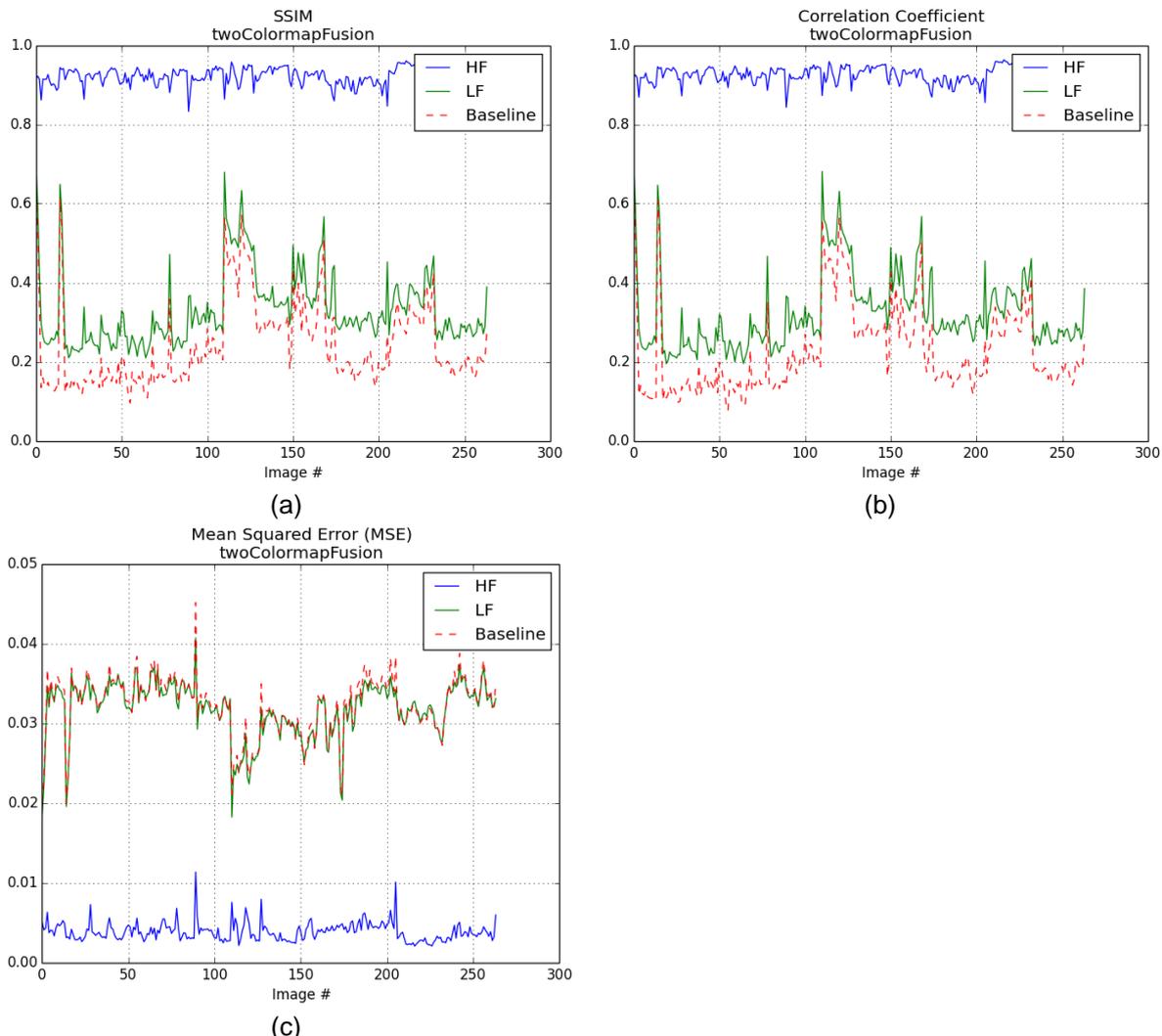

**Figure 7. (a), (b), (c) Images metrics of the dual colormap fusion algorithm for SSIM, correlation coefficient, and mean squared error respectively.**

All algorithms demonstrated increased information content in the fused image when compared to any single image. This is demonstrated in Figure 5, Figure 6, and Figure 7 subplots (a) and (b) by green and blue lines being greater than the dotted-red line, the baseline.

The CFAR saliency algorithm metric best preserved the relevant features in the HF image but has the worst performance for the LF imagery. The LF imagery features are best preserved with the dual colormap fusion algorithm as well as the SURF saliency algorithm with the SURF saliency algorithm having the better HF performance of the two. This result seems reasonable given the sophistication of the SURF saliency metric versus the two simpler methods based on a CFAR detector.

The results of the SSIM and NCC metrics are very similar. The SSIM is the product of three components[6] : mean, variance, and cross correlation. The dynamic range compression algorithm used to generate the input images into the fusion algorithm creates HF/LF images having similar means and variances. This causes the mean and variance terms in SSIM to cancel thus reducing the metrics to its cross correlation term.

## 6     CONCLUSION

We propose methods which ingest high frequency (HF) and low frequency (LF) SAS imagery and generate a color composite creating what we call a multiband SAS (MSAS) image.  The MSAS image contains the relevant portions of the HF and LF bands required by an analyst to interpret the scene and are defined using a spatial saliency metric computed for each band.  We then combine the saliency and acoustic backscatter measures to form the final MSAS image.

We investigate three fusion schemes. The first two schemes -- one based on a constant false alarm rate (CFAR) detector and one based on speeded up robust feature (SURF) densities -- fuse the data in a human visual system (HVS) focused color space CIELAB[1] while the third scheme fuses by using dual colormaps -- one for salient HF features and one for salient LF features.  We evaluate our results by examining three similarity metrics on the original images and the fused image. The metrics we examined are structural similarity index metric[3] (SSIM), normalized cross correlation (NCC), and mean-squared-error (MSE).

We find all three algorithms produce fused images containing information from both the HF and LF bands.  We also found the dual colormap and SURF saliency fusion algorithms give the best results based on the metrics we analysed.

Future work includes maturing the saliency algorithms perhaps basing them on points of interest relevant to an analyst specific  task, or focusing on a more HVS centric saliency metric such as Itti's model[7].